\documentclass{article}

\usepackage[final]{nips_2017}

\usepackage[utf8]{inputenc} 
\usepackage[T1]{fontenc}    
\usepackage{hyperref}       
\usepackage{url}            
\usepackage{booktabs}       
\usepackage{amsfonts}       
\usepackage{nicefrac}       
\usepackage{microtype}      

\usepackage{subcaption}
\usepackage{mathtools}

\usepackage{amsthm}
\usepackage{wrapfig}

\theoremstyle{definition}

\title{Transferring Agent Behaviors from Videos via Motion GANs}

\author{
Ashley D. Edwards \\
College of Computing\\
Georgia Institute of Technology\\
Atlanta, GA 30332 \\
\texttt{aedwards8@gatech.edu} \\
\And
Charles L. Isbell Jr. \\
College of Computing \\
Georgia Institute of Technology\\
Atlanta, GA 30332 \\
\texttt{isbell@cc.gatech.edu} \\
}

\begin{document}

\maketitle
\begin{abstract}
A major bottleneck for developing general reinforcement learning agents is determining rewards that will yield desirable behaviors under various circumstances. We introduce a general mechanism for automatically specifying meaningful behaviors from raw pixels. In particular, we train a generative adversarial network to produce short sub-goals represented through motion templates. We demonstrate that this approach generates visually meaningful behaviors in unknown environments with novel agents and describe how these motions can be used to train reinforcement learning agents. 
\end{abstract}

\section{Introduction}
Reinforcement Learning (RL) has been shown to be a successful approach for solving complex problems such as games and robotics tasks. Still, progress is often hindered because the learned policies do not generalize well to multiple agents and environments, resulting in a need for a new reward function each time the problem changes. Traditional approaches use hand-crafted rewards to specify goals, or for more complex tasks, one can demonstrate the desired behavior and train agents via imitation learning. These representations can successfully train behaviors for known agents, but typically require domain knowledge that does not generalize to unexplored environments.

We often have a broad expectation for how objects should move in the real-world, even if we have not seen them before. We do not expect pigs to sprout wings and fly into the air, or clouds to float down onto the earth's surface. Inanimate materials should not become animate, nor should rivers flow upstream. Our expectations are shaped by the events we have experienced in the world. 
Inspired by this, we aim to use deep learning to learn about how objects should move in artificial and real-world environments, and use this information to inform how agents should act. 

Rather than requiring task-specific engineering for each problem, our approach aims to learn a general representation of motion from videos that can then be used to visually generate desired behaviors in the agent's environment. We develop a generative model of common motions by training an image-to-image model~\cite{pix2pix2016} to compute motion templates~\cite{bobick2001recognition,davis1999recognizing} from still images within a wide array of environments. With this model, we can generate a plan of motion for agents and environments that are similar to those that were observed in the training set. We then aim to use this visual plan to learn a policy that imitates the desired behavior. 

In this initial work, we focus on developing the framework for predicting motions. We will demonstrate that the learned model can generate motions in environments that it trained on as well as in unfamiliar environments consisting of novel agents. 

\section{Background}
\label{sec:background}
\begin{figure}[h]
  \centering
  \begin{subfigure}{.325\linewidth}
    \centering
    \includegraphics[width=\linewidth]{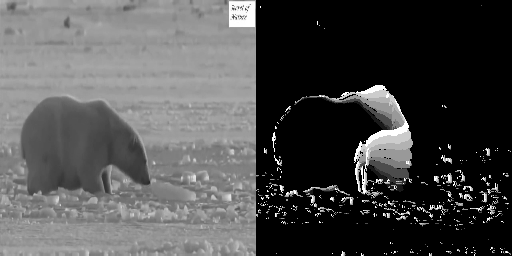}
  \end{subfigure}
  \begin{subfigure}{.325\linewidth}
    \centering
    \includegraphics[width=\linewidth]{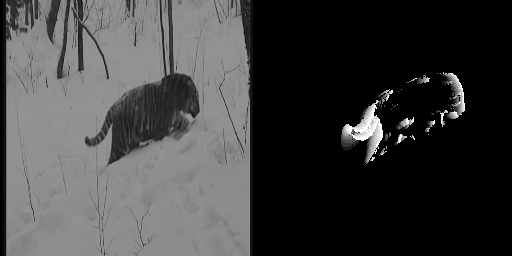}
  \end{subfigure}  
    \begin{subfigure}{.325\linewidth}
    \centering
    \includegraphics[width=\linewidth]{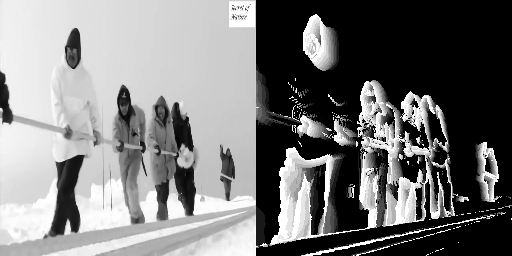}
  \end{subfigure}  
  \caption[Caption for motion template]{Motion templates computed from videos obtained from the WildLife Documentary (WLD) Dataset ~\cite{chen2017discover} of a polar bear, tiger, and group of people.}  
  \label{fig:motion}
\end{figure} 
We now briefly discuss methods we use in our approach: reinforcement learning, motion templates, and image-to-image translation.

\subsection{Reinforcement Learning}
Reinforcement Learning (RL) problems are described through a Markov Decision Process $\langle S, A, T, R \rangle$~\cite{suttonbarto}. The set $S$ consists of the states $s \in S$ in the environment. An agent takes actions $a \in A$ and receives rewards $r \in R(s)$ that specify the goals for the problem. The transition function $T(s, a, s')$ represents the probability that the agent will land in state $s'$ after taking action $a$ in state $s$. A policy $\pi(s,a)$ represents the probability of taking action $a$ in state $s$ and we typically aim to find policies that maximize the total long-term expected reward. 

An action-value, or Q-value, $Q(s, a)$ represents the expected discounted cumulative reward an agent will receive after taking action $a$ in state $s$, then following $\pi$ thereafter. We typically are interested in computing optimal Q-values:
$$
Q^*(s,a) = \max_{\pi} \mathbb{E}\Bigg[\sum_{k=0}^{\infty}\gamma^{k}r_{t+k+1} | s_t = s, a_t = a, \pi\Bigg]
$$
where $0 \le \gamma < 1$ is a discount factor that encodes how rewards retain their value over-time.

\subsection{Motion Templates}
In our approach, we represent goals as motion templates (see Figure~\ref{fig:motion}), which are 2D spatio-temporal representations of motion obtained from a sequence of images---typically from the segmented frames of a video~\cite{bobick2001recognition,davis1999recognizing}. Movement that occurred more recently in time has a higher pixel intensity in the template than earlier motion and depicts both where and when motion occurred. 

Calculating a motion template is an iterative process. The first step is to obtain a silhouette image of the motion that has occurred between each frame. The silhouette is computed by taking the absolute difference between two images and then computing the binary threshold, which sets all pixels below a threshold to $0$ and all pixels above the threshold to $1$.

A function $\Psi(\textbf{I})$ computes the motion template $\mu$ for a sequence of images $i_1, i_2, \dots, i_n \in \textbf{I}$. Let $\sigma_t$ represent a silhouette image at time $t$. To calculate the motion template $\mu$ of $\textbf{I}$, we first compute a silhouette image $\sigma_1, \sigma_2, \dots, \sigma_{n-1}$ between all consecutive images $(i_1, i_2), (i_2, i_3), \dots, (i_{n-1}, i_n)$. Then $\forall_{x,y}$, where $x$ and $y$ are respective column and row pixel locations, we can compute $\mu_{t,x,y}$ for time $t = 1, 2, \dots, n$:
\[
    \mu_{t,x,y} =
\begin{cases}
    \tau,& \text{if } \sigma_{t,x,y} > 0 \\
    0, & \text{else if }  \mu_{t-1, x, y} < (\tau - \delta) \\
    \mu_{t-1, x, y}, & \text{otherwise}
\end{cases}
\]
In words, the function increases the intensity of the pixel at $x, y$ if movement has occurred at the current iteration $t$. Here, $\delta$ and $\tau$ are both parameters that influence how much $\mu_t$ is decayed. The parameter $\tau$ is a representation for the current time in the sequence and increases as $t$ increases. The parameter $\delta$ represents the duration of the motion template and controls how quickly pixels decay. Essentially,  $\Psi(\textbf{I})$ layers the silhouette images and weights them by time.

\subsection{Image-to-image translation}
We use image-to-image translation---a recent approach that uses Generative Adversarial Networks (GANs) to translate one image into another~\cite{pix2pix2016}---to automatically generate motion templates. Image-to-image models have been used to convert images of edges to handbags and images of day to night. Given an input image, a generative model attempts to generate the translated output, while a discriminative model predicts whether the pair was obtained from the training data or generated by the GAN. The generative network is trained to fool the discriminator and to output images that match the ground truth, while the discriminative model is trained to make correct predictions.

\section{Related work}
\label{sec:related}
Goals in reinforcement learning are traditionally defined through rewards that indicate when desired states have been reached. A major benefit of RL is that good policies can often be learned from a single reward, but a consequence of this sparsity is that learning the policies can be slow and inefficient. In general, we argue that RL suffers from a~\emph{reward shortage}, particularly because new tasks often require engineering more rewards, even when the environment has been seen before. Representing goals separately from rewards offers more portability, for example by using target images~\cite{edwards2016perceptual,finn2017deep}, but this still requires specifying goals for each problem. An alternative is to learn policies that can generalize across environments. Transfer learning aims to port learned behaviors from one domain to another~\cite{taylor2009transfer}, for example by initializing the parameters of policies in unsolved tasks~\cite{ammar2015unsupervised}, or by transferring skills across untrained robots~\cite{devin2016learning}. Training in simulation and then transferring the knowledge to the real-world can often be more efficient than training there directly (e.g. ~\cite{rusu2016sim,sadeghi2016cad,tobin2017domain, tzeng2016adapting, zhang2015towards}). Such ``sim-to-real'' approaches tend to focus on transferring across separate realizations of similar domains. Learning from Demonstration (LfD) can also be used when specifying a goal is difficult, or when a problem is too challenging for an agent to solve on its own. Inverse RL aims to infer a reward function from expert demonstrations~\cite{abbeel2004apprenticeship} or policies can be learned directly from demonstrations~\cite{schaal1999imitation}. Many recent works have trained agents such as simulated and real robots from the raw pixels of videos.~\cite{edwards2016perceptual,sermanet2016unsupervised,sermanet2017time,liu2017imitation}. These approaches often focus on learning tasks from specific demonstrations for these problems. Our approach aims to learn general models from videos. In particular, we develop a hierarchy for specifying and satisfying goals, similar to works that develop temporally-extended sequences of actions known as options and controllers for selecting from them~\cite{sutton1999between,bacon2017option,kulkarni2016hierarchical,vezhnevets2017feudal}. Finally, many approaches have learned models of the world (see~\cite{polydoros2017survey} for a survey), but these methods often require access to the underlying MDP. Our approach learns a general model from pixels that can be applied to novel environments. 
\section{Approach}
Our approach aims to learn a general model of motion from videos. We use the recent image-to-image architecture to predict motion templates from still images. We call this representation Motion GAN (MoGAN). This paper focuses on how to train this model, but we will also discuss how we can use these generated motions to construct short-term policies for unknown agents within unobserved environments.

MoGANs are inspired by previous work for learning to act from motion templates~\cite{edwards2016perceptual}. In that approach, a simulated robot was trained by comparing motion templates of the robot to those of humans. Motion templates allow for a denser reward function that naturally shapes the desired behavior. The reward was specified as the similarity between the templates, but utilized hand-crafted features for comparability. We aim to automatically generate plans of motion that are appropriate for the current environment. In particular, we develop a hierarchy in which a meta-controller (MoGAN) generates a high-level plan of action in the form of motion and a controller (the policy) then learns actions that satisfy this goal. Developing the controller remains as future work.

\subsection{Meta-controller}
We now describe how we train the MoGAN model, or  meta-controller. Given a dataset of videos, we segment each video into a sequence of $n$ frames. Then we use each mini-sequence to compute a motion template. We construct frame-motion pairs by taking the initial frame of each sequence and pairing with the respective motion template. With these pairs, we can train the image-to-image model to predict motions from still images. We aim to use generated behaviors as goals for RL.

\subsection{Multimodal outputs}
One problem with image-to-image translation is the model is forced to learn a one-to-one mapping, when in fact there are many possible outputs of the network. To address this problem, we develop a multimodal network that allows for multiple outputs. We expect to find that the network will output~\emph{visual} options that indicate multiple behaviors. In this network, we still train each output to fool the discriminator, but now the generator is only penalized on the minimum distance between the ground truth and each of the generated outputs. 

\section{Experiments and results}
\begin{figure}[t]
  \centering
  \begin{subfigure}{.16\linewidth}
    \centering
    \includegraphics[width=\linewidth]{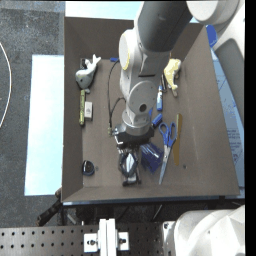}
  \end{subfigure}
    \begin{subfigure}{.16\linewidth}
    \centering
    \includegraphics[width=\linewidth]{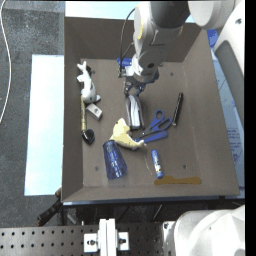}
  \end{subfigure}
  \begin{subfigure}{.16\linewidth}
    \centering
    \includegraphics[width=\linewidth]{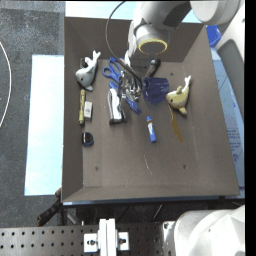}
  \end{subfigure}
  \begin{subfigure}{.16\linewidth}
    \centering
    \includegraphics[width=\linewidth]{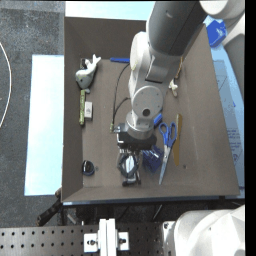}
  \end{subfigure}
    \begin{subfigure}{.16\linewidth}
    \centering
    \includegraphics[width=\linewidth]{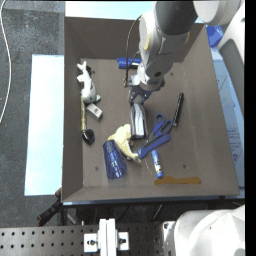}
  \end{subfigure}
  \\
  \begin{subfigure}{.16\linewidth}
    \centering
    \includegraphics[width=\linewidth]{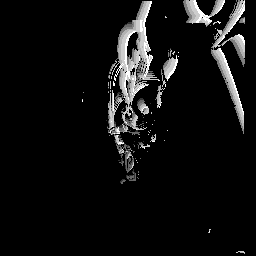}
  \end{subfigure}  
  \begin{subfigure}{.16\linewidth}
    \centering
    \includegraphics[width=\linewidth]{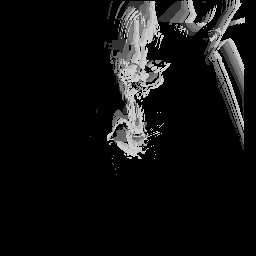}
  \end{subfigure} 
  \begin{subfigure}{.16\linewidth}
    \centering
    \includegraphics[width=\linewidth]{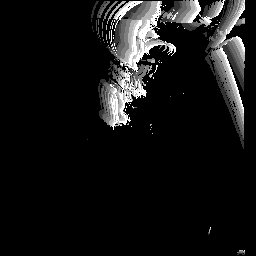}
  \end{subfigure} 
  \begin{subfigure}{.16\linewidth}
    \centering
    \includegraphics[width=\linewidth]{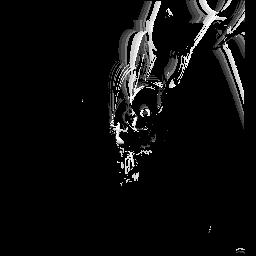}
  \end{subfigure} 
    \begin{subfigure}{.16\linewidth}
    \centering
    \includegraphics[width=\linewidth]{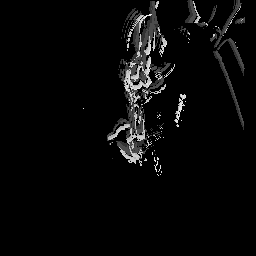}
  \end{subfigure} 
    \\
  \begin{subfigure}{.16\linewidth}
    \centering
    \includegraphics[width=\linewidth]{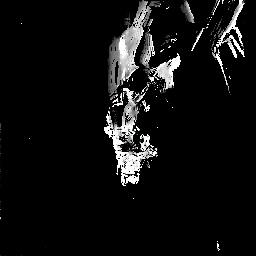}
  \end{subfigure}  
  \begin{subfigure}{.16\linewidth}
    \centering
    \includegraphics[width=\linewidth]{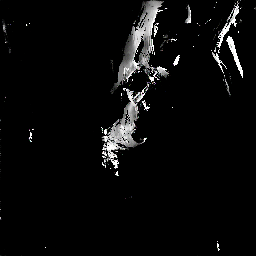}
  \end{subfigure} 
  \begin{subfigure}{.16\linewidth}
    \centering
    \includegraphics[width=\linewidth]{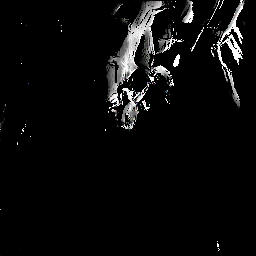}
  \end{subfigure} 
  \begin{subfigure}{.16\linewidth}
    \centering
    \includegraphics[width=\linewidth]{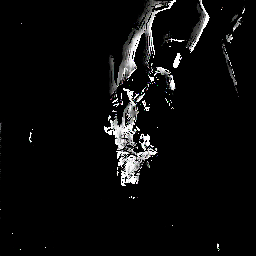}
  \end{subfigure} 
    \begin{subfigure}{.16\linewidth}
    \centering
    \includegraphics[width=\linewidth]{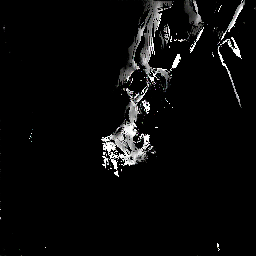}
  \end{subfigure} 
  \caption{Generated push motion templates on unseen test images. The top row represents the initial frame used to compute the motion templates. The middle row represents the ground truth motion template. The bottom row represents the generated motion template.}  
  \label{fig:pushes}
\end{figure} 
\begin{figure}[t]
  \centering
  \begin{subfigure}{.325\linewidth}
    \centering
    \includegraphics[width=\linewidth]{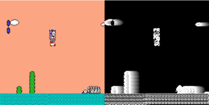}
  \end{subfigure}
    \begin{subfigure}{.325\linewidth}
    \centering
    \includegraphics[width=\linewidth]{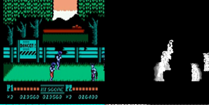}
  \end{subfigure}
  \begin{subfigure}{.325\linewidth}
    \centering
    \includegraphics[width=\linewidth]{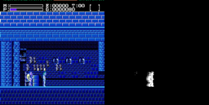}
  \end{subfigure}
  \\
 \vspace{.035cm}
  \begin{subfigure}{.325\linewidth}
    \centering
    \includegraphics[width=\linewidth]{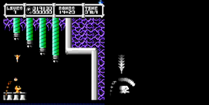}
  \end{subfigure}  
  \begin{subfigure}{.325\linewidth}
    \centering
    \includegraphics[width=\linewidth]{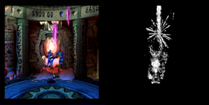}
  \end{subfigure} 
  \begin{subfigure}{.325\linewidth}
    \centering
    \includegraphics[width=\linewidth]{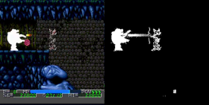}
  \end{subfigure} 
  \caption{Sample of frame-motion training pairs obtained from video games. The left image represents the initial frame and the right represents the computed motion template. }  
  \label{fig:train_video_games}
\end{figure} 
\begin{figure}[t]
  \centering
  \begin{subfigure}{.137\linewidth}
    \centering
    \includegraphics[width=\linewidth]{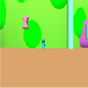}
  \end{subfigure}
    \begin{subfigure}{.137\linewidth}
    \centering
    \includegraphics[width=\linewidth]{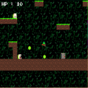}
  \end{subfigure}
  \begin{subfigure}{.137\linewidth}
    \centering
    \includegraphics[width=\linewidth]{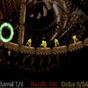}
  \end{subfigure}
  \begin{subfigure}{.137\linewidth}
    \centering
    \includegraphics[width=\linewidth]{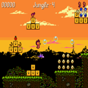}
  \end{subfigure}
    \begin{subfigure}{.137\linewidth}
    \centering
    \includegraphics[width=\linewidth]{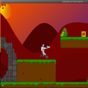}
  \end{subfigure}
      \begin{subfigure}{.137\linewidth}
    \centering
    \includegraphics[width=\linewidth]{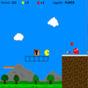}
  \end{subfigure}
        \begin{subfigure}{.137\linewidth}
    \centering
    \includegraphics[width=\linewidth]{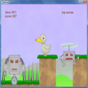}
  \end{subfigure}
  \\
  \begin{subfigure}{.137\linewidth}
    \centering
    \includegraphics[width=\linewidth]{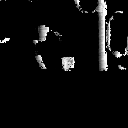}
  \end{subfigure}  
  \begin{subfigure}{.137\linewidth}
    \centering
    \includegraphics[width=\linewidth]{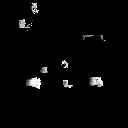}
  \end{subfigure} 
  \begin{subfigure}{.137\linewidth}
    \centering
    \includegraphics[width=\linewidth]{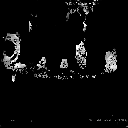}
  \end{subfigure} 
  \begin{subfigure}{.137\linewidth}
    \centering
    \includegraphics[width=\linewidth]{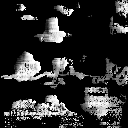}
  \end{subfigure} 
    \begin{subfigure}{.137\linewidth}
    \centering
    \includegraphics[width=\linewidth]{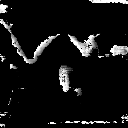}
  \end{subfigure} 
    \begin{subfigure}{.137\linewidth}
    \centering
    \includegraphics[width=\linewidth]{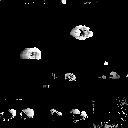}
  \end{subfigure} 
        \begin{subfigure}{.137\linewidth}
    \centering
    \includegraphics[width=\linewidth]{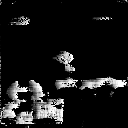}
  \end{subfigure}
  \caption{Generated motion templates from screenshots of unseen video games obtained from pygame.org. The top row represents the initial frame and the bottom row represents the generated motion template.}  
  \label{fig:video_games}
\end{figure} 
\begin{figure}[t]
  \centering
  \begin{subfigure}{.137\linewidth}
    \centering
    \includegraphics[width=\linewidth]{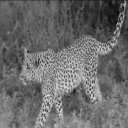}
  \end{subfigure}
    \begin{subfigure}{.137\linewidth}
    \centering
    \includegraphics[width=\linewidth]{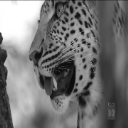}
  \end{subfigure}
  \begin{subfigure}{.137\linewidth}
    \centering
    \includegraphics[width=\linewidth]{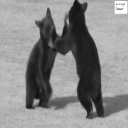}
  \end{subfigure}
  \begin{subfigure}{.137\linewidth}
    \centering
    \includegraphics[width=\linewidth]{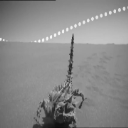}
  \end{subfigure}
    \begin{subfigure}{.137\linewidth}
    \centering
    \includegraphics[width=\linewidth]{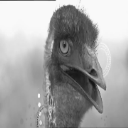}
  \end{subfigure}
      \begin{subfigure}{.137\linewidth}
    \centering
    \includegraphics[width=\linewidth]{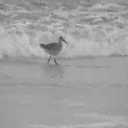}
  \end{subfigure}
        \begin{subfigure}{.137\linewidth}
    \centering
    \includegraphics[width=\linewidth]{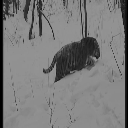}
  \end{subfigure}
  \\
  \begin{subfigure}{.137\linewidth}
    \centering
    \includegraphics[width=\linewidth]{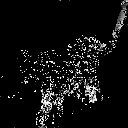}
  \end{subfigure}  
  \begin{subfigure}{.137\linewidth}
    \centering
    \includegraphics[width=\linewidth]{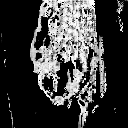}
  \end{subfigure} 
  \begin{subfigure}{.137\linewidth}
    \centering
    \includegraphics[width=\linewidth]{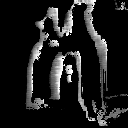}
  \end{subfigure} 
  \begin{subfigure}{.137\linewidth}
    \centering
    \includegraphics[width=\linewidth]{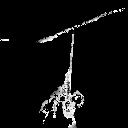}
  \end{subfigure} 
    \begin{subfigure}{.137\linewidth}
    \centering
    \includegraphics[width=\linewidth]{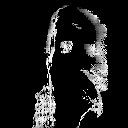}
  \end{subfigure} 
    \begin{subfigure}{.137\linewidth}
    \centering
    \includegraphics[width=\linewidth]{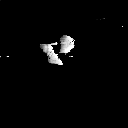}
  \end{subfigure} 
        \begin{subfigure}{.137\linewidth}
    \centering
    \includegraphics[width=\linewidth]{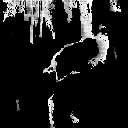}
  \end{subfigure}
  \caption{Generated motion templates for unseen real-world images from the WLD Dataset. The top row represents the initial frame and the bottom row represents the generated motion template.}  
  \label{fig:cheetah}
\end{figure} 
\begin{figure}[t]
  \centering
  \begin{subfigure}{.16\linewidth}
    \centering
    \includegraphics[width=\linewidth]{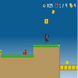}
  \end{subfigure}
    \begin{subfigure}{.16\linewidth}
    \centering
    \includegraphics[width=\linewidth]{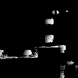}
  \end{subfigure}
  \begin{subfigure}{.16\linewidth}
    \centering
    \includegraphics[width=\linewidth]{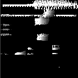}
  \end{subfigure}
    \begin{subfigure}{.16\linewidth}
    \centering
    \includegraphics[width=\linewidth]{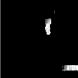}
  \end{subfigure}
  \\
  \begin{subfigure}{.16\linewidth}
    \centering
    \includegraphics[width=\linewidth]{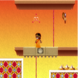}
  \end{subfigure}  
  \begin{subfigure}{.16\linewidth}
    \centering
    \includegraphics[width=\linewidth]{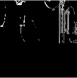}
  \end{subfigure} 
  \begin{subfigure}{.16\linewidth}
    \centering
    \includegraphics[width=\linewidth]{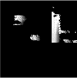}
  \end{subfigure} 
    \begin{subfigure}{.16\linewidth}
    \centering
    \includegraphics[width=\linewidth]{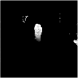}
  \end{subfigure} 
    \\
  \begin{subfigure}{.16\linewidth}
    \centering
    \includegraphics[width=\linewidth]{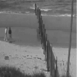}
  \end{subfigure}  
  \begin{subfigure}{.16\linewidth}
    \centering
    \includegraphics[width=\linewidth]{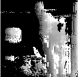}
  \end{subfigure} 
  \begin{subfigure}{.16\linewidth}
    \centering
    \includegraphics[width=\linewidth]{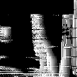}
  \end{subfigure} 
    \begin{subfigure}{.16\linewidth}
    \centering
    \includegraphics[width=\linewidth]{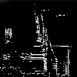}
  \end{subfigure} 
\caption{Learned visual options in video games from pygame.org and a real-world image of people walking~\cite{10.1109/TPAMI.2009.112}. The left represents the input, and each other column represents an output of the network.}
\end{figure} 
Our experiments aim to demonstrate that MoGANs can transfer behaviors from videos. For now, we focus on qualitative analysis of generated behaviors. 
\subsection{Intra-domain behaviors}
We first evaluate intra-domain predictions. We collected videos from Google's Push Dataset~\cite{finn2016unsupervised} and aimed to predict motions within this environment. This dataset contains videos of a robotic arm pushing objects within a bin. To obtain the training set, we segmented each video into $5$ frames and computed motion templates for each segment. The resulting training set consisted of 1,200 frame-motion pairs. Figure~\ref{fig:pushes} shows the results. It is clear that the model learns to predict movement of the arm. Additionally, while objects that are unlikely to move are ignored from the scene, objects that are close to the gripper are predicted to move. 
\subsection{Transferred behaviors}
The next experiments aimed to evaluate the generality of MoGANs. We trained MoGANs to generate motions from videos of video game play-throughs obtained from archive.org~\footnote{\url{https://archive.org/details/speed_runs}}. To obtain the training set, we segmented each video into $10$ frames and computed motion templates for each segment. The resulting training set consisted of 101,227 frame-motion pairs (see Figure~\ref{fig:train_video_games} for examples). We expected to find that we could predict motions of video games outside of this training set. We tested on unseen platform games from pygame.org. We show results in Figure~\ref{fig:video_games}. We again found that the model was able to segment salient objects from the screen. The model predicted that the background would remain stationary while the player and other objects would move from the left to right, which is standard behavior for platformers. 

Interestingly, we also found that MoGAN learned to detect salient objects in the real world, even though it was trained only on video games, as shown in Figure~\ref{fig:cheetah}. One possible reason is that in order to detect motion in video games, MoGAN must first learn to attend to important regions in the scene, and then predict what the motions will be. Another reason could be that the domain randomization forces the model to be robust to environment changes, which is an argument for why some sim-to-real approaches have been successful~\cite{tobin2017domain}. These results demonstrate the robustness of MoGANs and additionally suggest that we can use the model for attention in RL problems.  
\subsection{Multimodal behaviors}
Finally, we evaluate performance for multimodal outputs. We indeed observed that the model learned visual option-like behavior, as one output predicted moving to the left, one to the right, and one jumping in the air. 
\section{Conclusion and future work}
In this paper, we have shown how behaviors can be generated for both seen and unseen environments. We introduced MoGAN, a meta-controller for producing goal motions from still images. We should point out that our approach does suffer from a common problem with GANs, as the model sometimes predicts meaningless behaviors. We aim to use recent techniques for improving the quality of the outputs to improve the model. Future work will entail using the generated motions to plan with reinforcement learning. We now outline a plan for this approach.

\subsection{Controller}
We now describe how we can train an RL agent. We assume that the agent's environment consists of visual inputs. Given an observation, we aim to use the learned meta-controller to generate the expected behavior. In particular, every $n$ steps, the meta-controller generates a new goal given the current observation. We additionally use the last $n$ frames from the current episode to~\emph{compute} a motion template. We represent the agent's state as its current observation, the generated goal, and its current motion template. As such, it should learn to plan based on its current goal.

\subsection{Reward Function}
We aim to use two components for the reward function. First, we can compute the similarity between the generated and computed motion templates by extracting features from the discriminative network and taking the $L1$ distance between the templates. We can additionally use the discriminator to determine if the agent's computed motion template looks like a motion that should occur naturally. The generated motion template acts as a visual plan for meaningful behaviors, while the discriminator aims to discourage unnatural ones. This mechanism for training is motivated by Generative Adversarial Imitation Learning (GAIL)~\cite{ho2016generative}, which aims to generate behaviors that fool a discriminator into predicting if behaviors were executed by the demonstrator. 

\small
\bibliographystyle{abbrv}
\bibliography{references}

\end{document}